\documentclass[10pt,twocolumn,letterpaper]{article}

\usepackage[pagenumbers]{iccv} % To force page numbers, e.g. for an arXiv version

\usepackage{amsfonts}
\usepackage{amsmath}
\usepackage{amssymb}

\usepackage{makecell}
\usepackage{multicol}
\usepackage{multirow}
\usepackage{fontawesome}
\usepackage{pifont}

\usepackage[table,dvipsnames]{xcolor}

\definecolor{crafter}{RGB}{208,159,18}

\usepackage[pagebackref=false,breaklinks,colorlinks,allcolors=crafter]{hyperref}

\hypersetup{
  colorlinks=true,
  linkcolor=crafter,
  citecolor=crafter,
  urlcolor=crafter,
}

\usepackage{listings}
\usepackage{tcolorbox}
\usepackage{framed}
\usepackage{fontawesome}

\title{
Learning to Generate 4D LiDAR Sequences
}

\author{
Ao Liang$^{1,2,3}$\quad 
Youquan Liu$^{4}$\quad 
Yu Yang$^{5}$\quad 
Dongyue Lu$^{1}$\quad 
Linfeng Li$^{1}$\\
Lingdong Kong$^{1,*}$\quad 
Huaici Zhao$^{3,\dagger}$\quad 
Wei Tsang Ooi$^{1,\dagger}$
\\[0.4ex]
{\small$^1$NUS}
~~ 
{\small$^2$UCAS}
~~
{\small$^3$SIA, CAS}
~~
{\small$^4$FDU}
~~
{\small$^5$ZJU}
\\[0.4ex]
\faGlobe~\textbf{Project Page:} \href{https://lidarcrafter.github.io}{\textcolor{crafter}{\textcolor{crafter}{{\textbf{\textsl{Link}}}}}} 
~\quad~
\faGithubAlt~\textbf{GitHub:} \href{https://github.com/worldbench/lidarcrafter}{\textcolor{crafter}{{\textbf{\textsl{Link}}}}}
~\quad~
\faGear~\textbf{EvalKit:} \href{https://lidarcrafter.github.io}{\textcolor{crafter}{\textcolor{crafter}{{\textbf{\textsl{Link}}}}}}
}

\def\paperID{11}
\def\confName{ICCV}
\def\confYear{2025}

\begin{document}

% \maketitle
\twocolumn[{
    \renewcommand\twocolumn[1][]{#1}
    \maketitle
    \begin{center}
    \vspace{-0.35cm}
    \includegraphics[width=\textwidth]{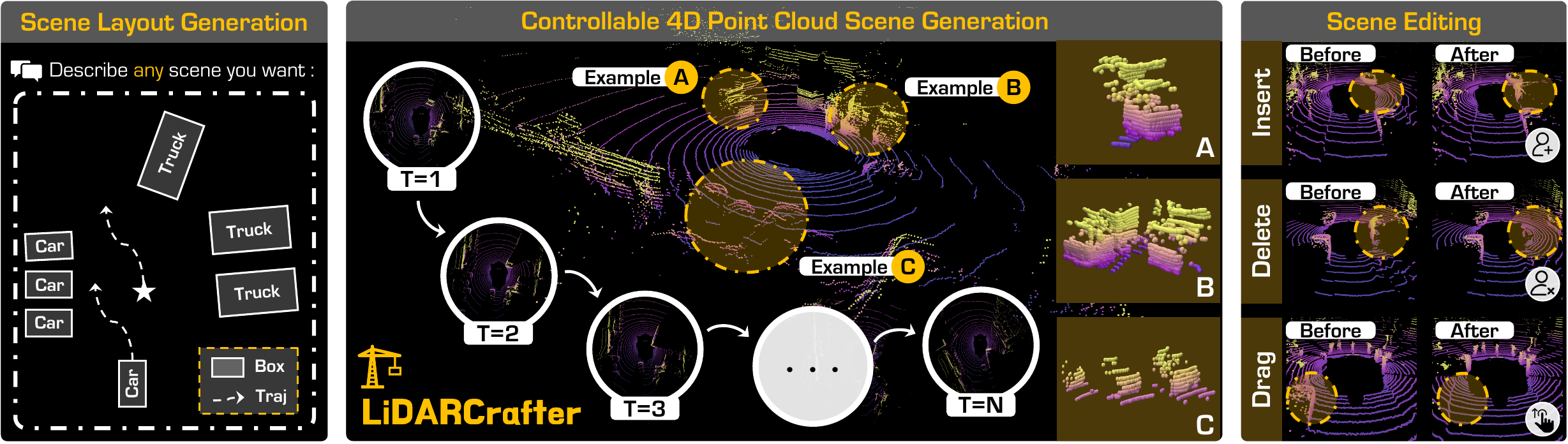}
    \vspace{-0.65cm}
    \captionof{figure}{This work introduces \textbf{LiDARCrafter}, a LiDAR-based 4D generative world model that supports controllable point cloud layout generation (\textbf{left}), dynamic sequential scene generation (\textbf{center}), and rich scene editing applications (\textbf{right}).}
    \label{fig:teaser}
    \end{center}
}]
 
% Main Body
\begin{abstract}
While generative world models have advanced video and occupancy-based data synthesis, LiDAR generation remains underexplored despite its importance for accurate 3D perception. Extending generation to 4D LiDAR data introduces challenges in controllability, temporal stability, and evaluation. We present \textbf{LiDARCrafter}, a unified framework that converts free-form language into editable LiDAR sequences. Instructions are parsed into ego-centric scene graphs, which a tri-branch diffusion model transforms into object layouts, trajectories, and shapes. A range-image diffusion model generates the initial scan, and an autoregressive module extends it into a temporally coherent sequence. The explicit layout design further supports object-level editing, such as insertion or relocation. To enable fair assessment, we provide \textbf{EvalSuite}, a benchmark spanning scene-, object-, and sequence-level metrics. On nuScenes, LiDARCrafter achieves state-of-the-art fidelity, controllability, and temporal consistency, offering a foundation for LiDAR-based simulation and data augmentation.
\end{abstract}
 
\vspace{-0.2cm}
\section{Introduction}
\label{sec:intro}

Generative world models are reshaping autonomous driving by enabling scalable simulation and interpretation of sensor-rich environments~\cite{survey_3d_4d_world_models,hu2023gaia,mei2024dreamforge}. Recent advances have focused on structured modalities such as videos and occupancy grids, whose dense and regular representations align naturally with image or voxel pipelines. Video-based methods \cite{hu2023gaia,mei2024dreamforge,zhang2025epona} leverage autoregression and richer conditioning, while BEV-based approaches such as MagicDrive~\cite{gao2023magicdrive} enforce temporal consistency across frames~\cite{wang2024drivedreamer,zhao2025drivedreamer}. Occupancy-based works \cite{zheng2024occworld,wang2024occsora,zuo2025gaussianworld,bian2024dynamiccity}, capture fine spatial structure for downstream tasks. Multimodal frameworks \cite{li2025uniscene,guo2025genesis} further align cross-modal signals for consistency. Despite this progress, LiDAR, a core modality for precise 3D geometry and all-weather robustness, remains comparatively underexplored.

\begin{figure*}[t]
    \centering
    \includegraphics[width=0.99\textwidth]{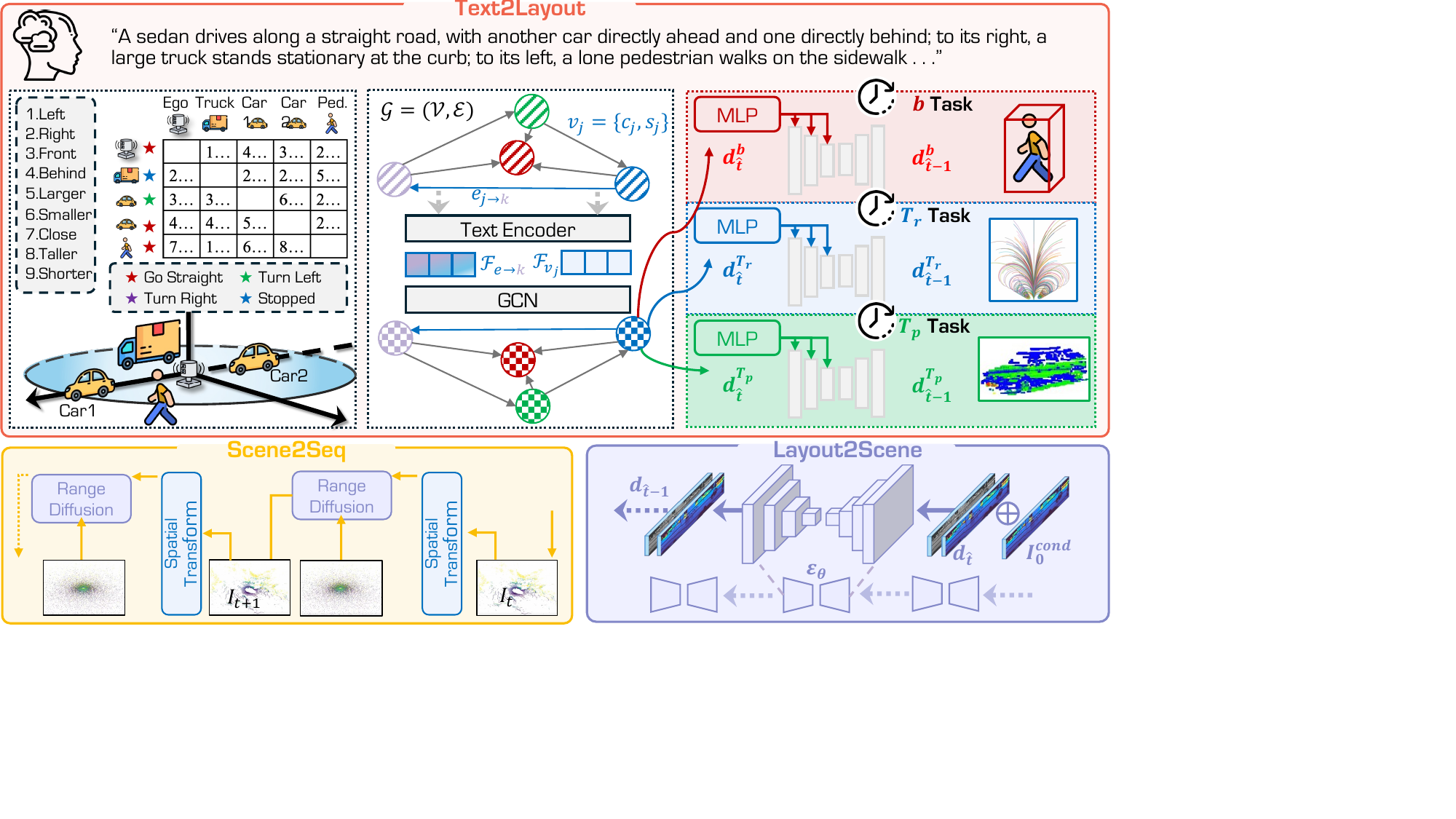}
    \vspace{-0.2cm}
    \caption{
    \textbf{Overview of the LiDARCrafter framework.} In the \textbf{Text2Layout} stage (\textit{cf.}~Sec.~\ref{sec:layout_generation}), the natural-language instruction is parsed into an ego-centric scene graph, and a tri-branch diffusion network generates 4D conditions for bounding boxes, future trajectories, and object point clouds. In the \textbf{Layout2Scene} stage (\textit{cf.}~Sec.~\ref{sec:static_pointcloud_generation}), a range-image diffusion model uses these conditions to generate a static LiDAR frame. In the \textbf{Scene2Seq} stage (\textit{cf.}~Sec.~\ref{sec:4D_pointcloud_generation}), an autoregressive module warps historical points with ego and object motion priors to generate each subsequent frame, producing a temporally coherent LiDAR sequence.
    }
    \label{fig:framework}
\end{figure*}

LiDAR point clouds present unique challenges. They are sparse, irregular, and unordered~\cite{kong2023robo3d,liang2025pi3det,liu2023seal,xu2024superflow,zhu2025spiral,liu2025lalalidar,liu2025veila}, making direct application of image- or voxel-based techniques ineffective. Early efforts, such as LiDARGen~\cite{zyrianov2022learning} project 360° scans to range images and adapt pixel-based diffusion, while subsequent works like RangeLDM~\cite{hu2024rangeldm}, R2DM~\cite{nakashima2024lidar}, and R2Flow~\cite{nakashima2024fast} improved single-frame fidelity. Other methods such as Text2LiDAR~\cite{wu2024text2lidar}, WeatherGen~\cite{wu2025weathergen}, and UltraLiDAR~\cite{xiong2023ultralidar} introduced diverse conditioning or editing capabilities. Yet, most are restricted to static scans or lack temporal modeling, leaving 4D sequence generation and fine-grained control unresolved.

A central obstacle is \textbf{spatial controllability}. Existing models often require costly inputs such as HD maps~\cite{swerdlow2024street} or 3D bounding boxes~\cite{zhang2024perldiff,yang2024drivearena}, while text-only methods~\cite{hu2023gaia,wu2024text2lidar} are more accessible but lack spatial precision. Indoor scene synthesis has addressed this trade-off with intermediate scene graphs~\cite{zhai2024echoscene,yang2025mmgdreamer}, but such strategies are not yet established for outdoor, dynamic LiDAR streams. Beyond controllability, LiDAR world models also lack \textbf{temporal coherence}: single-frame synthesis cannot capture occlusions or motion patterns, and naïve cross-frame attention overlooks the geometric continuity of point clouds. Finally, unlike video models that benefit from benchmarks such as VBench~\cite{huang2024vbench}, LiDAR has no standardized protocols to evaluate scene-, object-, and sequence-level quality.

We introduce \textbf{LiDARCrafter}, the first unified framework for controllable 4D LiDAR sequence generation. At its core is an explicit, object-centric 4D layout that bridges free-form language instructions with LiDAR geometry and motion. In the \textbf{Text2Layout} stage, a large language model parses descriptions into an ego-centric scene graph, which a tri-branch diffusion network expands into object boxes, trajectories, and shapes. \textbf{Layout2Scene} converts this layout into a high-fidelity initial scan using a range-image diffusion backbone, enabling precise editing such as object insertion or relocation. \textbf{Scene2Seq} autoregressively extends the sequence by warping foreground and background points with motion priors to maintain long-term temporal consistency. To close the evaluation gap, we release \textbf{EvalSuite}, the first benchmark that jointly scores semantic correctness, layout validity, and sequence smoothness.

In summary, LiDARCrafter establishes a new paradigm for LiDAR-based world modeling by combining intuitive language-driven control, explicit layout conditioning, and temporally stable generation. It offers both a practical synthesis tool and a standardized benchmark, providing the community with a foundation for controllable and consistent 4D LiDAR simulation.
\section{LiDARCrafter: 4D LiDAR World Model}
\label{sec:method}

We introduce \textbf{LiDARCrafter}, the first generative world model dedicated to LiDAR, which transforms free-form instructions into temporally coherent 4D point cloud sequences with object-level control. The core idea is to maintain an explicit 4D layout that bridges language descriptions with LiDAR geometry. As shown in Fig.~\ref{fig:framework}, the framework follows three stages: \textbf{Text2Layout}, which lifts language into a structured 4D layout; \textbf{Layout2Scene}, which generates a controllable first scan; and \textbf{Scene2Seq}, which autoregressively extends the sequence. We further introduce \textbf{EvalSuite}, a protocol for standardized evaluations.

\begin{figure*}[t]
    \centering
    \includegraphics[width=\textwidth]{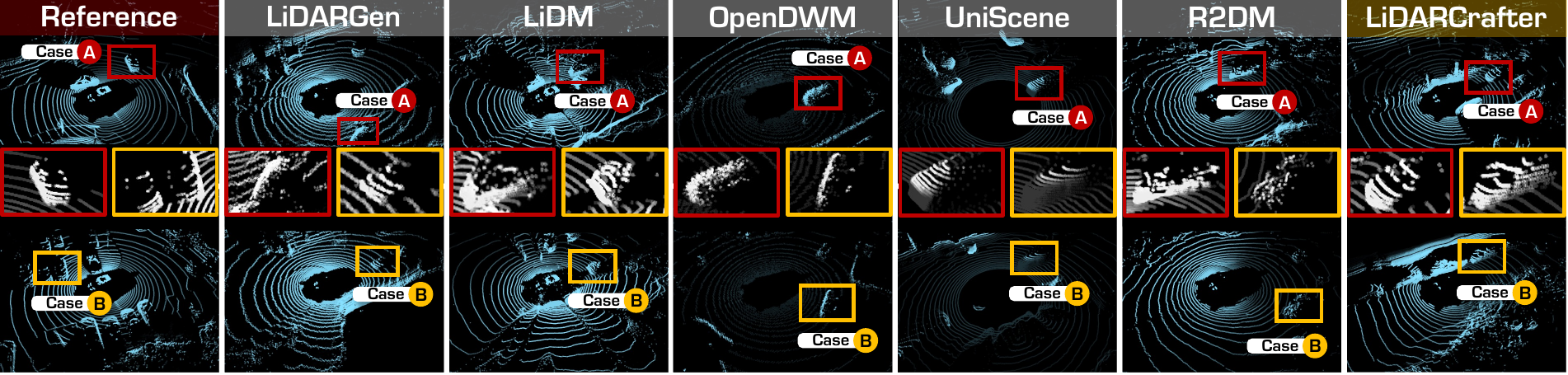}
    \vspace{-0.65cm}
    \caption{
    Visual comparisons of \textbf{single-frame LiDAR point cloud generation} on nuScenes \cite{caesar2020nuscenes}. LiDARCrafter produces the pattern closest to the ground truth, with notably superior foreground quality compared to other methods. Best viewed at high resolution.
    }
    \label{fig:scene_comparison}
\vspace{-0.3cm}
\end{figure*}

\begin{figure}[t]
    \centering
    \includegraphics[width=\linewidth]{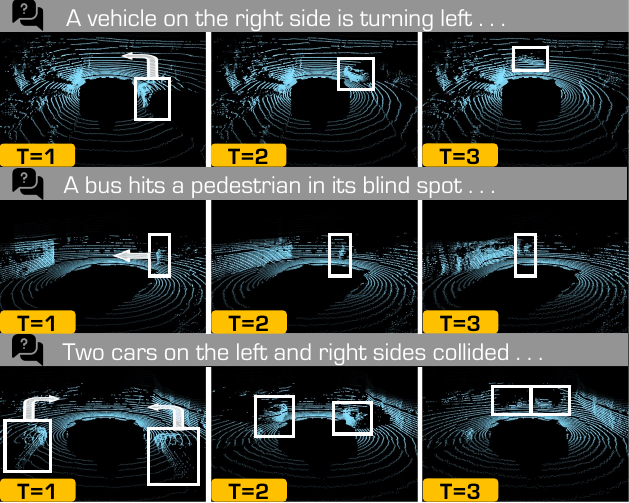}
    \vspace{-0.65cm}
    \caption{
    \textbf{Diverse corner cases} generated by LiDARCrafter with object-centric 3D controllability. Best viewed at high resolution. Frames are arranged sequentially from left to right.
    }
    \label{fig:corner_case}
\end{figure}

\subsection{Text2Layout: Language to 4D Layout}
\label{sec:layout_generation}

Since natural language lacks the spatial precision needed for LiDAR synthesis, we construct an intermediate scene graph. An LLM parses the user prompt into an ego-centric graph $\mathcal{G}=(\mathcal{V},\mathcal{E})$, where nodes represent the ego vehicle and dynamic objects with semantic labels and motion states, while edges encode spatial relations. This explicit graph captures both semantic and relational cues.

Each node is lifted into a layout tuple $\mathcal{O}_i=(\mathbf{b}_i,\boldsymbol{\delta}_i,\mathbf{p}_i)$, including a 3D bounding box, a trajectory of planar offsets, and canonical shape points. To enrich these with contextual semantics, we process the graph with a TripletGCN~\cite{johnson2018image}, embedding nodes and edges with a frozen CLIP encoder~\cite{radford2021learning}. The resulting features condition a tri-branch diffusion decoder: one branch denoises boxes, another predicts trajectories, and a third synthesizes coarse shapes. This structured 4D layout provides the foundation for later LiDAR synthesis while supporting explicit object-level control.

\subsection{Layout2Scene: Controlled LiDAR Generation}
\label{sec:static_pointcloud_generation}

Given the 4D layout, LiDARCrafter generates the initial LiDAR frame using a range-image diffusion model, which preserves LiDAR geometry while leveraging efficient convolutional backbones~\cite{kong2023rangeformer,nakashima2024lidar}. To address sparsity, particularly for small or distant objects, we condition the network on compact object representations encoding category, pose, and shape priors, projected onto the range view~\cite{kirby2024logen}. A lightweight attention layer propagates context across objects, and scene-level embeddings provide global conditioning. During denoising, the noisy map is combined with this conditioning to yield a coherent first scan.

The explicit layout also enables fine-grained editing. By modifying layout tuples (\eg, inserting, deleting, or dragging objects), users can re-synthesize scenes with only local changes, preserving the rest of the scan. This makes LiDARCrafter suitable for interactive scenario design in simulation and planning research.

\subsection{Scene2Seq: Autoregressive Sequence Synthesis}
\label{sec:4D_pointcloud_generation}

To extend a single scan into a full 4D sequence, we adopt an autoregressive strategy. Unlike video, where textures change every frame, LiDAR scenes are largely static except for moving agents and ego motion. LiDARCrafter exploits this by warping background points with the ego pose and foreground objects with their predicted trajectories. This warp provides a strong geometric prior at each timestep, which the diffusion model refines into a clean range map. To prevent accumulated drift, we include a warp from the first frame to every later frame, ensuring long-term stability. The result is a temporally consistent sequence where motion and occlusion patterns remain realistic.

\subsection{EvalSuite: Comprehensive Evaluation}
\label{sec:evaluation}

LiDAR generation requires metrics beyond static fidelity. Our \textbf{EvalSuite} measures quality across three levels:  
\textbf{Scene-level}: evaluating global realism and distributional fidelity of entire scenes.
\textbf{Object-level}: verifying semantic correctness, bounding box geometry, and detection confidence.  
\textbf{Temporal-level}: assessing motion smoothness and frame-to-frame transform accuracy.  

Together, these metrics provide the first standardized benchmark for 4D LiDAR sequence generation, enabling fair and holistic evaluation of future methods.

\section{Experiments}
\label{sec:experiments}

\begin{table}[t]
    \centering
    \caption{Evaluations of \textbf{scene-level fidelity} for LiDAR generation on the \textit{nuScenes} dataset. MMD values are reported in $10^{-4}$ and JSD in $10^{-2}$. Lower is better for all metrics ($\downarrow$).} 
    \vspace{-0.3cm}
    \resizebox{\linewidth}{!}{
    \begin{tabular}{c|r|r|cc|cc|cc}
    \toprule
    \multirow{2}{*}{\textbf{\#}} & \multirow{2}{*}{\textbf{Method}} & \multirow{2}{*}{\textbf{Venue}} & \multicolumn{2}{c|}{\textbf{Range}} & \multicolumn{2}{c|}{\textbf{Points}} & \multicolumn{2}{c}{\textbf{BEV}} 
    \\
    & & & \textbf{FRD}$\downarrow$ & \textbf{MMD}$\downarrow$ & \textbf{FPD}$\downarrow$ & \textbf{MMD}$\downarrow$ & \textbf{JSD}$\downarrow$ & \textbf{MMD}$\downarrow$ 
    \\
    \midrule\midrule
    \multirow{3.5}{*}{\rotatebox{90}{\textbf{Voxel}}}
    &UniScene & CVPR'25 &      –   &   –  & $976.47$ & $29.06$ & $31.55$ & $13.61$ \\
    &OpenDWM       & CVPR'25  &      –   &   –  & $714.19$ & $21.95$ & $20.17$ & $5.61$ \\
    &OpenDWM-DiT   & CVPR'25  &      –   &   –  & $381.91$ & $12.46$ & $19.90$ & $5.73$ \\
    \midrule
    \multirow{5.5}{*}{\rotatebox{90}{\textbf{Range}}}    
    &LiDARGen & ECCV'22  & $759.65$ & $1.71$ & $159.35$ & $35.52$ & $5.74$ & $2.39$ \\
    &LiDM          & CVPR'24  &  $495.54$ & $0.18$ & $210.20$ &  $8.45$ & $5.86$ & $0.73$ \\
    &RangeLDM      & ECCV'24  &      –   &   –  &     –   &    –   & $5.47$ & $1.92$ \\
    &R2DM          & ICRA'24  &  $243.35$ & $1.40$ &  $33.97$ &  $1.62$ & $3.51$ & $0.71$ \\
    \cmidrule{2-9}
    &\textbf{LiDARCrafter} & \textbf{Ours} & \cellcolor{crafter!15}$\mathbf{194.37}$ & \cellcolor{crafter!15}$\mathbf{0.08}$  & \cellcolor{crafter!15}$\mathbf{8.64}$ & \cellcolor{crafter!15}$\mathbf{0.90}$ & \cellcolor{crafter!15}$\mathbf{3.11}$ & \cellcolor{crafter!15}$\mathbf{0.42}$ 
    \\
    \bottomrule
    \end{tabular}}
    \label{tab:scene_point_generation}
    \vspace{-0.1cm}
\end{table}
\begin{table}[t]
    \centering
    \caption{
    Comparison of \textbf{foreground object quality} using FDC ($\uparrow$), which reflects detector confidence on generated scenes. \#Box is the average number of boxes per frame.
    }
    \vspace{-0.3cm}
    \resizebox{\linewidth}{!}{
    \begin{tabular}{c|r|r|cccc|c}
    \toprule
    \textbf{\#} & \textbf{Method} & \textbf{Venue} & \textbf{Car}$\uparrow$ & \textbf{Ped}$\uparrow$ & \textbf{Truck}$\uparrow$ & \textbf{Bus}$\uparrow$ 
      & \textbf{\#Box} \\
    \midrule\midrule
    \multirow{3.1}{*}{\rotatebox{90}{\small \textbf{Uncond.}}}
    &LiDARGen      & ECCV'22  
      & $$0.57$$ & $$0.29$$ & $$0.42$$ & $$0.38$$ 
      & $$0.364$$ \\
    &LiDM          & CVPR'24  
      & $$0.65$$ & $$0.22$$ & $$0.45$$ & $$0.31$$ 
      & $$0.28$$  \\
    &R2DM          & ICRA'24  
      & $$0.54$$ & $$0.29$$ & $$0.39$$ & $$0.35$$ 
      & $$0.53$$  \\
    \midrule
    \multirow{4.5}{*}{\rotatebox{90}{\textbf{Cond.}}}
    &UniScene      & CVPR'25  
      & $$0.53$$ & $$0.28$$ & $$0.35$$ & $$0.25$$ 
      & $$0.98$$  \\
    &OpenDWM       & CVPR'25  
      & $$0.74$$ & $$0.30$$ & $$0.51$$ & $$0.44$$ 
      & $$0.54$$  \\
    &OpenDWM-DiT   & CVPR'25  
      & $$0.78$$ & $$0.32$$ & $\mathbf{0.56}$ & $$0.51$$ 
      & $$0.64$$  \\
    \cmidrule{2-8}
    &\textbf{LiDARCrafter} & \textbf{Ours} & \cellcolor{crafter!15}$\mathbf{0.83}$ & \cellcolor{crafter!15}$\mathbf{0.34}$ & \cellcolor{crafter!15}$0.55$ & \cellcolor{crafter!15}$\mathbf{0.54}$ & \cellcolor{crafter!15}$\mathbf{1.84}$ 
    \\
    \bottomrule
    \end{tabular}}
    \label{tab:fdc}
    \vspace{-0.1cm}
\end{table}

\begin{table}[t]
    \centering
    \caption{
    Evaluation of \textbf{object-level fidelity} for LiDAR generation. MMD is reported in $10^{-4}$, and JSD in $10^{-2}$.
    }
    \vspace{-0.3cm}
    \resizebox{\linewidth}{!}{
    \begin{tabular}{c|r|r|cccc}
    \toprule
    \textbf{\#}& \textbf{Method}       & \textbf{Venue} 
      & \textbf{FPD}$\downarrow$ 
      & \textbf{P-MMD}$\downarrow$ 
      & \textbf{JSD}$\downarrow$ 
      & \textbf{MMD}$\downarrow$ \\
    \midrule\midrule
    \multirow{3.1}{*}{\rotatebox{90}{\small \textbf{Uncond.}}}
    &LiDARGen      & ECCV'22 
      & $$1.39$$  & $$0.15$$ 
      & $$0.20$$  & $$16.22$$ \\
    &LiDM          & CVPR'24 
      & $$1.41$$  & $$0.15$$ 
      & $$0.19$$  & $$13.49$$ \\
    &R2DM          & ICRA'24 
      & $$1.40$$  & $$0.15$$ 
      & $$0.17$$  & $$12.76$$ \\
    \midrule
    \multirow{4.5}{*}{\rotatebox{90}{\textbf{Cond.}}}
    &UniScene      & CVPR'25 
      & $$1.19$$  & $$0.18$$ 
      & $$0.23$$  & $$16.65$$ \\
    &OpenDWM       & CVPR'25 
      & $$1.49$$  & $$0.19$$ 
      & $$0.16$$  & $$9.11$$  \\
    &OpenDWM-DiT   & CVPR'25 
      & $$1.48$$  & $$0.18$$ 
      & $\mathbf{0.15}$  & $$9.02$$  \\
    \cmidrule{2-7}
    &\textbf{LiDARCrafter} & \textbf{Ours} 
    & \cellcolor{crafter!15}$\mathbf{1.03}$ & \cellcolor{crafter!15}$\mathbf{0.13}$ & \cellcolor{crafter!15}$\mathbf{0.15}$ & \cellcolor{crafter!15}$\mathbf{5.48}$ 
    \\
    \bottomrule
    \end{tabular}}
    \label{tab:object_metrics}
    \vspace{-0.3cm}
\end{table}

\begin{table}[t]
    \centering
    \caption{
    Comparison of \textbf{temporal consistency} in 4D LiDAR generation. TTCE measures transformation error from point cloud registration; CTC computes Chamfer Distance between adjacent frames. Numbers indicate frame intervals.}
    \vspace{-0.3cm}
    \resizebox{\linewidth}{!}{
    \begin{tabular}{r|r|cc|cccc}
    \toprule
    \multirow{2}{*}{\textbf{Method}} & \multirow{2}{*}{\textbf{Venue}} 
      & \multicolumn{2}{c|}{\textbf{TTCE}$\downarrow$} 
      & \multicolumn{4}{c}{\textbf{CTC}$\downarrow$} \\
    & 
      & \textbf{3} & \textbf{4} 
      & \textbf{1} & \textbf{2} & \textbf{3} & \textbf{4} \\
    \midrule\midrule
    UniScene      & CVPR'25  
      & $$2.74$$ & $$3.69$$ 
      & $$0.90$$ & $$1.84$$ & $$3.64$$ & $\mathbf{3.90}$ \\
    OpenDWM       & CVPR'25  
      & $$2.68$$ & $$3.65$$ 
      & $$1.02$$ & $$2.02$$ & $$3.37$$ & $$5.05$$ \\
    OpenDWM-DiT   & CVPR'25  
      & $$2.71$$ & $$3.66$$ 
      & $\mathbf{0.89}$ & $\mathbf{1.79}$ & $$3.06$$ & $$4.64$$ \\
    \midrule
    \textbf{LiDARCrafter} & \textbf{Ours} & \cellcolor{crafter!15}$\mathbf{2.65}$ & \cellcolor{crafter!15}$\mathbf{3.56}$ & \cellcolor{crafter!15}$1.12$ & \cellcolor{crafter!15}$2.38$ & \cellcolor{crafter!15}$\mathbf{3.02}$ & \cellcolor{crafter!15}$4.81$ 
    \\
    \bottomrule
    \end{tabular}}
    \label{tab:sequence_coherence}
    \vspace{-0.4cm}
\end{table}

We evaluate \textbf{LiDARCrafter} on nuScenes~\cite{caesar2020nuscenes} using both classical LiDAR generation metrics (FRD, FPD, JSD, MMD) and our EvalSuite for measuring object-, layout- and sequence-level generation quality. Comparisons are made against recent LiDAR generative models, including R2DM \cite{nakashima2024lidar}, R2Flow \cite{nakashima2024fast}, and OpenDWM \cite{opendwm}.

\subsection{Scene-Level Generation}
At the scene level, we assess both whole-scan fidelity and the accuracy of synthesized foregrounds. As shown in \cref{tab:scene_point_generation}, LiDARCrafter attains the lowest FRD and FPD, outperforming prior methods by a notable margin. Qualitative comparisons in \cref{fig:scene_comparison} confirm that our model reconstructs realistic global structure while preserving sharp object geometry, whereas alternatives often suffer from noise or blurred backgrounds.

Foreground quality is further validated using a pre-trained VoxelRCNN detector~\cite{deng2021voxel}. We report Foreground Detection Confidence (FDC). LiDARCrafter achieves the highest scores across most categories (\cref{tab:fdc}), demonstrating that generated objects align closely with ground-truth semantics and boxes.

\subsection{Object-Level Generation}
At the object level, we benchmark fidelity and consistency under box-level conditioning. Using 2,000 car instances, LiDARCrafter achieves the best scores in FPD and MMD (\cref{tab:object_metrics}), showing superior reconstruction of fine-grained geometry compared to OpenDWM and R2Flow. 

% To assess semantic and geometric consistency, we introduce CFCA (classification accuracy) and CFSC (conditional shape consistency). A PointMLP classifier confirms strong category alignment with a CFCA of 73.5\%, and regression-based IoU evaluation indicates that our generated shapes remain geometrically consistent with target boxes (\cref{tab:cfca_cfsc}). Together, these results highlight that LiDARCrafter produces objects that are both semantically faithful and structurally coherent.

\subsection{Temporal-Level Generation}
We evaluate temporal consistency in 4D LiDAR generation in~\Cref{tab:sequence_coherence}. Temporal Transformation
Consistency Error (TTCE) measures the error between the predicted and ground-truth transformation matrices obtained via point cloud registration, while Chamfer Temporal Consistency (CTC) computes the Chamfer Distance between consecutive frames. Our approach achieves the lowest TTCE scores across both frame intervals and maintains competitive CTC performance at all intervals, demonstrating strong temporal coherence.
\section{Conclusion}
\label{sec:conclusion}
We presented \textbf{LiDARCrafter}, a unified framework for controllable 4D LiDAR sequence generation and editing. By leveraging scene graph descriptors, the multi-branch diffusion model, and an autoregressive generation strategy, our approach achieves fine-grained controllability and strong temporal consistency. Experiments on nuScenes demonstrate clear improvements over existing methods in fidelity, coherence, and controllability. Beyond high-quality data synthesis, LiDARCrafter enables the creation of safety-critical scenarios for robust evaluation of downstream autonomous driving systems. Future work will explore multi-modal extensions and further efficiency improvements.
% References
\clearpage
\clearpage
{
    \small
    \bibliographystyle{ieeenat_fullname}
    \bibliography{main}
}

\end{document}